\documentclass{article}
\usepackage{spconf,amsmath,epsfig}
\usepackage{amssymb}
\usepackage{todonotes}
\usepackage{multirow}

\title{Multi-modal Vision Transformers for Crop Mapping from Satellite Image Time Series}
\name{Theresa Follath{\normalfont\textsuperscript{1}}, David Mickisch{\normalfont\textsuperscript{1}}, Jan Hemmerling{\normalfont\textsuperscript{2}},
\textit{Stefan Erasmi}{\normalfont\textsuperscript{2}}, \textit{Marcel Schwieder}{\normalfont\textsuperscript{2}} \textit{and Beg\"{u}m Demir}{\normalfont\textsuperscript{1,3}}}
\address{\textsuperscript{1}Faculty of Electrical Engineering and Computer Science, Technische Universit\"{a}t Berlin, Germany\\
\textsuperscript{2}Th\"{u}nen Institute of Farm Economics\\
\textsuperscript{3}BIFOLD - Berlin Institute for the Foundations of Learning and Data, Germany}
\begin{document}
\maketitle
\begin{abstract}
Using images acquired by different satellite sensors has shown to improve classification performance in the framework of crop mapping from satellite image time series (SITS).
Existing state-of-the-art architectures use self-attention mechanisms to process the temporal dimension and convolutions for the spatial dimension of SITS.
Motivated by the success of purely attention-based architectures in crop mapping from single-modal SITS, we introduce several multi-modal multi-temporal transformer-based architectures. 
Specifically, we investigate the effectiveness of Early Fusion, Cross Attention Fusion and Synchronized Class Token Fusion within the Temporo-Spatial Vision Transformer (TSViT). 
Experimental results demonstrate significant improvements over state-of-the-art architectures with both convolutional and self-attention components.
\end{abstract}
\section{Introduction}
\label{sec:intro}
The availability of spatially explicit data on agricultural land-use is a pre-requisite for many monitoring, reporting and evaluation activities at different spatial scales from regional to global. Spatial data are frequently available at aggregated levels (e.g. districts) but knowledge on annual land-use at the field level is sparse. Satellite data of the recent Earth Observation missions can close this gap.  
The increased availability, complexity and amount of high-quality satellite images has triggered the development of algorithms for processing huge archives of diverse data structures. In particular, deep learning architectures have proven to be superior to traditional machine learning methods for environmental mapping from dense satellite image time series (SITS).
Initial studies on deep learning based approaches use convolutional neural networks (CNN) with either a one-dimensional kernel to process the temporal dimension on pixel level \cite{pelletier2019temporal} or a three-dimensional kernel \cite{ji20183d} to extract spatio-temporal features.
Other works rely on recurrent neural networks (RNN) with long short-term memory (LSTM) cells to extract temporal features \cite{russwurm2017temporal}. 
With the rise of transformer architectures, self-attention is used more and more for feature extraction from the temporal dimension, often in combination with convolutional layers for spatial feature extraction \cite{garnot2021panoptic}. 
However, the recently introduced purely transformer-based Temporo-Spatial Vision Transformer (TSViT) has shown superior classification performance compared to these types of methods \cite{tarasiou2023vits}.
The model achieves this with two transformer encoders: i) one for temporal features and ii) one for spatial features. Thereby, it applies self-attention only and disposes convolutional layers altogether.

The previous methods are designed for single modality SITS and applied exclusively to optical data. However, in \cite{garnot2022multi} it is shown that the joint use of multispectral data and complementary synthetic-aperture radar (SAR) data can boost classification performance for crop mapping. The authors apply an architecture initially developed for optical data to multi-modal data by Early Fusion. 
The model consists of an U-Net to process the spatial dimension of the image at each time point separately. In the deepest layer of the U-Net, self-attention is used to compute weights and collapse the temporal dimension of each pixel of the feature map. 
A pre-trained transformer-based method for crop mapping from multi-modal SITS is introduced in \cite{tseng2023lightweight}. It processes each pixel time series separately without taking the spatial dimension into account. 

All of the above mentioned methods provide good classification accuracies. However, they either apply convolutions to extract spatial features or do not process the spatial dimension at all.

The success of the purely self-attention-based TSViT in crop mapping from single-modal SITS makes it a promising architecture that can be  extended by multi-modal data fusion to further improve classification performance. 
In this paper, we investigate modifications to the TSViT architecture that enable it to operate on multi-modal SITS. 
Particularly, we study the effectiveness of Early Fusion (EF), Cross Attention Fusion (CAF) and Synchronized Class Token Fusion (SCTF) in the context of the TSViT. This leads to three different multi-modal TSViT (MM TSViT) architectures.

\section{Proposed Multi-modal Fusion Methods}
The proposed multi-modal fusion architectures operate on sets of co-registered SITS, $\{X_j \}_{j=1}^M$, where $M$ is the number of the different sensor modalities that we are considering.
An SITS, $X_j$, is defined as $X_j \in \mathbb{R}^{T_j \times H_j \times W_j \times C_j}$ where $T_j, H_j, W_j$ and $C_j$ refer to the number of time steps, height, width and number of channels of the $j$th modality, respectively. The co-registered SITS of $\{ X_j \}_{j=1}^M$ are associated to the same geographic area.
A pixel of an SITS is defined as $X_j(h,w) \in \mathbb{R}^{T_j \times C_j}$ with $h \in [H_j]$ and $w \in [W_j]$ being the position of the pixel within the spatial dimension of the SITS.

In the following, crop mapping is considered as a pixel-based classification task. Therefore, we assume for each set of co-registered SITS, a label map, $Y \in [0,1]^{H \times W \times K}$, where $K$ denotes the number of labels for the classification task. Each label map together with the corresponding SITS from different modalities are associated to the same geographical area. Since we assume a single label per pixel, $Y(h, w)$ is a one-hot vector with only a single non-zero entry. We further assume that the spatial resolution of the label map corresponds to the finest spatial resolution among the input modalities, i.e. $W = \max\{W_j\}_{j=1}^M$ and $H = \max\{H_j\}_{j=1}^M$.

The task of pixel-based classification can then be defined as finding a function that takes a set of co-registered SITS and generates an output, $\hat{Y} \in [ 0, 1 ]^{H \times W \times K}$, which approximates $Y$ as closely as possible.

In order to learn a joint feature representation from all input modalities, we develop three multi-modal architectures based on the TSViT \cite{tarasiou2023vits} which operates as follows on single-modal data:
Let $X \in \mathbb{R}^{T\times H \times W \times C}$ be an SITS from a single modality. First, it is divided into non-overlapping 3D-patches $x_p \in \mathbb{R}^{t \times h \times w \times C}$ for $p \in [N_T\cdot N_H \cdot N_W]$ with $N_T = \lfloor \frac{T}{t}\rfloor$, $N_H = \lfloor \frac{H}{h}\rfloor$ and $N_W = \lfloor \frac{W}{w}\rfloor$. Subsequently, they are flattened, projected to the token dimension $d$ and reshaped to $Z_T \in \mathbb{R}^{N_W\cdot N_H \times N_T \times d}$.
A temporal position embedding $P_T[t,:]\in \mathbb{R}^{N_T \times d}$ based on the times of acquisition is added. 
Afterwards, $K$ class tokens $Z_{Tcls} \in \mathbb{R}^{K\times d}$ are prepended to each sequence which results in $Z_T^0 = concat(Z_{Tcls}, Z_T + P_T[t,:])$ which are then passed through the first transformer encoder, computing temporal attention. 
Among the tokens in the resulting sequences, only the class tokens are kept. 
The first two dimensions are transposed to obtain $Z_S \in \mathbb{R}^{K\times N_H\cdot N_W \times d}$ and passed through a second transformer encoder, computing spatial attention.
The final class tokens are then projected from the token dimension back to their initial spatial dimensions and rearranged to return a class probability map, $\hat{Y} \in [0, 1]^{H \times W \times K}$.

In the following we describe our proposed multi-modal fusion architectures. When a single data modality is considered, single modality TSViT (denoted as SM TSViT) \cite{tarasiou2023vits} is used.

\subsection{Early Fusion (EF)}
In the EF architecture, the SITS of all modalities are stacked in the channel dimension before passing the data to the model. This requires co-registered SITS of equal spatial and temporal resolution, i.e. $W_j = W_k, H_j=H_k, T_j = T_k$ for all $j,k \in [M]$, which can be achieved by resampling. Then, a set of co-registered SITS $\{X_j\}_{j=1}^M$ is concatenated in the channel dimension:
\begin{align}
    \bar{X} = concat(X_1,\dots, X_M),
\end{align}
with $\bar{X} \in \mathbb{R}^{T\times H \times W \times \sum_{j=1}^MC_j}$.
This tensor is passed through an SM TSViT model as described above, returning the map of class probabilities, $\hat{Y} \in [0, 1]^{H\times W \times K}$.
By fusing the modalities as early as possible, we ensure that all components of the architecture can contribute to the extraction of the most relevant features of each modality.

\subsection{Synchronized Class Token Fusion (SCTF)}
\begin{figure}[]
    \centering	
    \includegraphics[width=0.8\linewidth]{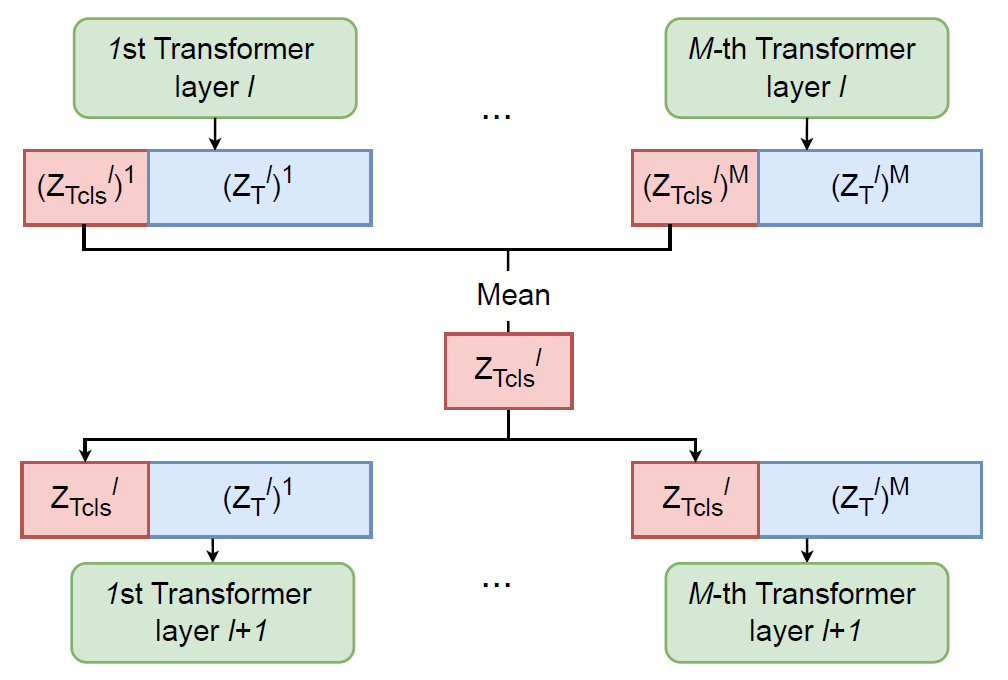}\\
    \caption{Illustration of the proposed Synchronized Class Token Fusion between layers $l$ and $l+1$ of the $M$ Encoders.}
    \label{fig:sctfusion}
\end{figure}

The SCTF architecture employs one transformer encoder for each modality between which class tokens are synchronized after each layer \cite{hoffmann2023transformer} (see Figure \ref{fig:sctfusion}).
For a set of co-registered SITS, each modality is processed separately until the first transformer encoder. This results in a set of $M$ input sequences $\{(Z^0_T)_j\}_{j=1}^M$. 
Each sequence is processed by a modality-specific transformer encoder with $L$ layers.
After a layer, $l \in [L]$, the set of modality-specific class tokens, $\{(Z_{Tcls}^l)_j\}_{j=1}^M$, is extracted. 
Instead of concatenating the tokens of each class and projecting back to the token dimension $d$, we take the mean of the $M$ class tokens for each class $k \in [K]$:
\begin{align}
     (Z_{Tcls}^l)^k = \frac{1}{M} \sum_{j=1}^M (Z_{Tcls}^l)_j^k.
\end{align}

The resulting modality-agnostic class tokens are prepended to the output feature tokens of the $l$th layer of all modality-specific encoders. 
These are then passed to layer $l+1$. 
After the final layer $L$, the mean aggregated class tokens $Z_{Tcls}^L$ are processed further by reshaping to $Z_S$ and passing to the second transformer encoder, as in the SM TSViT.

Synchronizing the class tokens after each layer allows for a controlled exchange of information between the modality-specific encoders. By allowing only the class tokens to interact, the exchange is limited to features that are relevant for the classification which constitute the most valuable information.

\subsection{Cross Attention Fusion (CAF)}
\begin{figure}[]
    \centering	
    \includegraphics[width=0.9\linewidth]{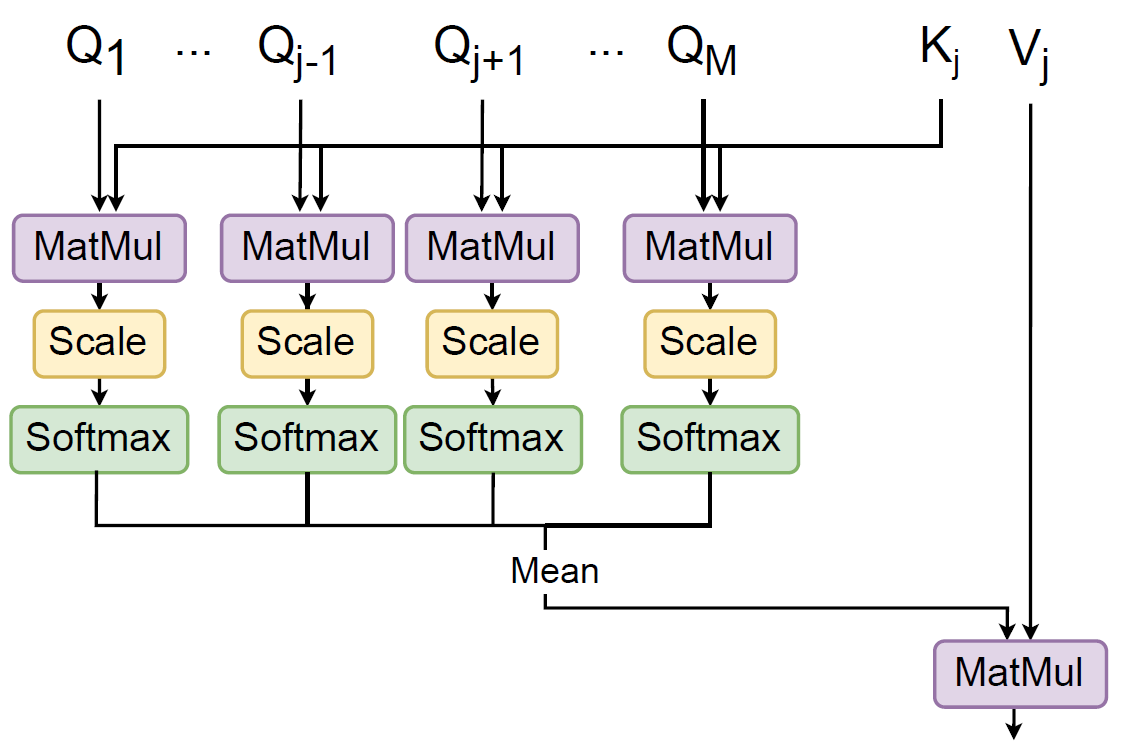}\\
    \caption{Illustration of Scaled Dot-Product Cross Attention for the $j$th modality-specific encoder. Queries $Q_1,\dots, Q_M$ are derived from $M$ different modalities.}
    \label{fig:crossattention}
\end{figure}
In the CAF architecture, the modalities are tokenized separately and passed through their own modality-specific temporal encoder. 
The encoders interact by exchanging information within the attention mechanism. The standard self-attention is defined as:
\begin{align}
    A(Q,K,V) &= softmax\bigg(\frac{QK^T}{\sqrt{d}} \bigg) V,
\end{align}
where $Q,K,V$ are query, key and value matrices respectively \cite{vaswani2017attention}.
In CAF, the queries are exchanged between modality-specific encoders in each encoder layer as shown in Figure \ref{fig:crossattention}. 
The specific encoders for modality $j\in [M]$ compute $Q_j, K_j, V_j$ from the tokens of the respective modality. 
For each pair of different modalities $i$ and $j$, we then compute the \textit{cross-attention weights} between $i$ and $j$: 
\begin{align}
a_{i, j} := softmax\bigg(\frac{Q_i K_j^T}{\sqrt{d}} \bigg).
\end{align}
Then, we compute cross-attention $A_j$ of the $j$th encoder by taking the mean of the cross-attention weights between modality $j$ and all other modalities and multiplying the resulting matrix with $V_j$ as in the self-attention case:
\begin{align}
A_j(\{ Q_{i} \}_{i=1}^M, K_j, V_j ) &:= \bigg(\frac{1}{M-1}\sum_{i \in [M], i \neq j} a_{i, j}\bigg) V_j.
\end{align}

After passing the input through the modality-specific temporal encoders, we extract the class tokens from the final layers of each encoder, aggregate their information by taking the class-wise mean and pass them to a single spatial transformer encoder that processes the input as the SM TSViT.

\section{Experimental results}
In this section, we briefly introduce the dataset on which we evaluate our architectures. Afterwards, we present our experimental setup and results.
\begin{table*}[]
    \centering
    \caption{Results (in \%) in Mean Accuracy (MA), Overall Accuracy (OA) and mean intersection-over-union (mIoU) achieved by U-TAE EF, SM TSViT considering S1, S2 and PF data separately and our MM TSViT architectures.}
    \vspace{4pt}
    \begin{tabular}{l|c|ccc|ccc}
        \hline
        \multirow{2}*{Metrics} & \multirow{2}*{U-TAE EF \cite{garnot2022multi}} & \multicolumn{3}{c|}{SM TSViT} & \multicolumn{3}{c}{Our MM TSViT} \\
        & & S1 & S2 & PF & EF & SCTF & CAF \\
        \hline
        MA & 67.70 & 65.45 & 75.12 & 72.57 & \textbf{80.34} & 79.72 & 79.38 \\
        OA & 89.75 & 85.07 & 88.72 & 88.79 & 89.79 & \textbf{90.50} & 90.39 \\
        mIoU & 55.68 & 52.43 & 61.08 & 60.30 & 66.96 & \textbf{68.39} & 66.66  
    \end{tabular}
    \label{tab:results}
\end{table*}
\subsection{Dataset Description}
We evaluated our architectures on the EOekoLand dataset that encompasses three modalities: Sentinel-1 (S1), Sentinel-2 (S2), and Planet Fusion (PF) data. The dataset includes time series data for the years 2020, 2021 and 2022 and spans over two 24 km by 24 km test regions in Germany, one area in the north-east (northern Brandenburg) and one area in the south-east (central Bavaria) of Germany. All modalities in the dataset are synthesised equidistant time series. 

For the construction of the S2 time series all available imagery within the specified years with a cloud cover of less than 75\% were obtained.
We used the FORCE-framework for radiometrical correction and cloud masking, including atmospheric and topographic, BRDF and adjacency effect correction \cite{frantz2019FORCE}.
The 20 m resolution bands of S2 were spatially enhanced to a 10 m resolution. To create capless time series data, we employed a Radial Basis Function (RBF) filter ensemble \cite{schwieder2016rbf} with four kernels: $\sigma_1$, $\sigma_2$, $\sigma_3$ and $\sigma_4$. 
These kernels incorporate observations within ±11, ±23, ±63, and ±127 days of the target date, respectively, with a preference for data points closer to the target day. The large temporal range of $\sigma_4$ allowed synthesizing continuous observations in data-poor scenarios, especially winter months.

We also obtained all available ascending Sentinel-1A and Sentinel-1B GRD scenes for the target years and 
processed them to $\gamma$0 backscatter, including  calibration, radiometric flattening and terrain correction. We then separately synthesized the VH and VV polarized data into uniform 10-day intervals time series reflecting the methodology and settings used for the S2 data \cite{force-sar}.
Therefore, consistency in the temporal influence of the original data on the synthesized output is maintained.
The PF \mbox{data \cite{planetdatasheet}} has a synthetic daily temporal resolution at a spatial resolution of 3m.

Reference data for crop types were derived from Integrated Administration and Control System (IACS) vector data, which we categorized into 15 semantic classes.
 Areas not covered by the reference data were designated as a background class. For one test region, reference data limitations restricted our analysis to the years 2020 and 2021.

\subsection{Results}
To perform multi-modal fusion of all three modalities, the SITS from S1 and S2 were upscaled from \mbox{10 m} to the spatial resolution of the PF data of \mbox{3 m} by bilinear interpolation. 
Moreover, we select the PF images acquired at the dates that correspond to the 10 day interval of the interpolated Sentinel images.
We obtain the results of the SM TSViT on S1, S2 and PF SITS individually to serve as reference baselines.
Furthermore, we compare the performance of our methods to the U-TAE with Early Fusion (U-TAE EF) as introduced in \cite{garnot2022multi} with their provided hyperparameters.
For all TSViT architectures, we use a patch size of 2, a temporal encoder depth of 6, a spatial encoder depth of 2 and token dimension $d$ of 128. 
All models are trained with batch size 8 for 50 epochs employing an Adam optimizer for which we set the initial learning rate to $10^{-4}$. Furthermore, we apply random flipping as a data augmentation technique.
Mean Accuracy (MA), Overall Accuracy (OA) and mean intersection-over-union (mIoU) are used as metrics for evaluation.
The results are shown in Table \ref{tab:results}.

From the results, one can see that our MM TSViT architectures provide significant improvements compared to SM TSViT. This shows that utilizing data from multiple sensors indeed improves crop mapping performance. 
Moreover, it can be seen that the proposed architectures outperform the U-TAE EF.
In detail, all proposed architectures achieve 12\% or higher MA and an improved mIoU of at least 11\%. The OA of our architectures is only about 1\% higher than that of the U-TAE EF.

Among the proposed architectures, SCTF outperforms CAF in all metrics by a small margin. 
With EF on the other hand a slightly higher MA is achieved. 
However, it should be noted that utilizing a separate temporal encoder for each modality in SCTF and CAF increases computational complexity of the model, whereas it is only marginally increased with EF.

\section{Conclusion}
In this paper we have studied the effectiveness of Early Fusion, Synchronized Class Token Fusion and Cross Attention Fusion in the context of the multi-modal TSViT. 
In contrast to existing architectures, these are, to the best of our knowledge, the first multi-modal fusion architectures for crop mapping from SITS that process both the temporal and the spatial dimensions with attention mechanisms without incorporating any convolutional or recurrent modules. 
Experimental results show that our architectures outperform the baselines by a large margin. 
The advantages of the proposed architectures are highlighted by the comparison to a state-of-the-art model that employs convolutional and self-attention modules with Early Fusion.
All these results prove the superiority of our purely attention-based architectures for crop mapping from multi-modal SITS.
As future work we plan to explore potentially more lightweight fusion architectures.

\section{Acknowledgement}

This work is supported by the German Ministry for Economic Affairs and Climate Action through the EOekoLand Project under Grant 50RP2230B. The Authors would like to thank Dr. Judith Br\"uggemann from FiBL for the valuable discussions on the design of the dataset.

\bibliographystyle{IEEEbib}
\bibliography{strings,refs}

\begin{thebibliography}{10}

\bibitem{pelletier2019temporal}
C.~Pelletier, G.~I. Webb, and F.~Petitjean,
\newblock ``Temporal convolutional neural network for the classification of satellite image time series,''
\newblock {\em Remote Sensing}, vol. 11, no. 5, pp. 523, 2019.

\bibitem{ji20183d}
S.~Ji, C.~Zhang, A.~Xu, Y.~Shi, and Y.~Duan,
\newblock ``3{D} convolutional neural networks for crop classification with multi-temporal remote sensing images,''
\newblock {\em Remote Sensing}, vol. 10, no. 1, pp. 75, 2018.

\bibitem{russwurm2017temporal}
M.~Ru{\ss}wurm and M.~Korner,
\newblock ``Temporal vegetation modelling using long short-term memory networks for crop identification from medium-resolution multi-spectral satellite images,''
\newblock {\em Proceedings of the IEEE/CVF Conference on Computer Vision and Pattern Recognition Workshops}, 2017, pp. 11--19.

\bibitem{garnot2021panoptic}
V.~S.~F. Garnot and L.~Landrieu,
\newblock ``Panoptic segmentation of satellite image time series with convolutional temporal attention networks,''
\newblock {\em Proceedings of the IEEE/CVF International Conference on Computer Vision}, 2021, pp. 4872--4881.

\bibitem{tarasiou2023vits}
M.~Tarasiou, E.~Chavez, and S.~Zafeiriou,
\newblock ``Vi{T}s for {SITS}: Vision transformers for satellite image time series,''
\newblock {\em Proceedings of the IEEE/CVF Conference on Computer Vision and Pattern Recognition}, 2023, pp. 10418--10428.

\bibitem{garnot2022multi}
V.~S.~F. Garnot, L.~Landrieu, and N.~Chehata,
\newblock ``Multi-modal temporal attention models for crop mapping from satellite time series,''
\newblock {\em ISPRS Journal of Photogrammetry and Remote Sensing}, vol. 187, pp. 294--305, 2022.

\bibitem{tseng2023lightweight}
G.~Tseng, I.~Zvonkov, M.~Purohit, D.~Rolnick, and H.~Kerner,
\newblock ``Lightweight, pre-trained transformers for remote sensing timeseries,''
\newblock {\em arXiv preprint arXiv:2304.14065}, 2023.

\bibitem{hoffmann2023transformer}
D.~S. Hoffmann, K.~N. Clasen, and B.~Demir,
\newblock ``Transformer-based multi-modal learning for multi-label remote sensing image classification,''
\newblock {\em IEEE International Geoscience and Remote Sensing Symposium}, 2023, pp. 4891--4894.

\bibitem{vaswani2017attention}
A.~Vaswani, N.~Shazeer, N.~Parmar, J.~Uszkoreit, L.~Jones, A.~N. Gomez, L.~Kaiser, and I.~Polosukhin,
\newblock ``Attention is all you need,''
\newblock {\em Advances in neural information processing systems}, vol. 30, 2017.

\bibitem{frantz2019FORCE}
D.~Frantz,
\newblock ``Force - {L}andsat+{S}entinel-2 analysis ready data and beyond,''
\newblock {\em Remote Sensing}, vol. 11, 2019.

\bibitem{schwieder2016rbf}
M.~Schwieder, P.~J. Leitão, M.~M. {da Cunha Bustamante}, L.~G. Ferreira, A.~Rabe, and P.~Hostert,
\newblock ``Mapping {B}razilian savanna vegetation gradients with {L}andsat time series,''
\newblock {\em International Journal of Applied Earth Observation and Geoinformation}, vol. 52, pp. 361--370, 2016.

\bibitem{force-sar}
F.~Lobert,
\newblock ``Force-sar framework,'' 2023,
\newblock url: https://github.com/felixlobert/force-sar/tree/main.

\bibitem{planetdatasheet}
Planet,
\newblock ``Planet fusion monitoring technical specification,'' 2021,
\newblock url: {https://assets.planet.com/docs/FusionTech-Spec\_v1.0.0.pdf}.

\end{thebibliography}

\end{document}